\documentclass[12pt, a4paper]{article}
\usepackage[font=footnotesize]{caption}
\usepackage[utf8]{inputenc}
\usepackage{graphicx}
\usepackage{authblk}
\usepackage{booktabs}
\usepackage{verbatim}
\usepackage{url}
\usepackage{caption}
\usepackage{subcaption}
\usepackage{indentfirst}
\usepackage{url}
\usepackage{amssymb}
\usepackage{amsmath}
\title{Multilyaer Multiset Neuronal Networks -- MMNNs}
\author{Alexandre Benatti$^1$ and Luciano da F. Costa$^2$}

\affil{
$^1$Institute of Mathematics and Statistics - DCC \\
University of S\~ao Paulo \\
Rua do Mat\~ao, 1010, S\~ao Paulo, SP 05508-090 Brazil 
\\ \vspace{0.5cm}
$^2$S\~ao Carlos Institute of Physics - DFCM \\
University of S\~ao Paulo \\
Av. Trabalhador S\~ao-carlense, 400, S\~ao Carlos, SP 13566-590 Brazil
}

\date{\emph{21th Jul., 2023}}

\begin{document}

\maketitle

\begin{abstract}
The coincidence similarity index, based on a combination of the Jaccard and overlap similarity indices, has noticeable properties in comparing and classifying data, including enhanced selectivity and sensitivity, intrinsic normalization, and robustness to data perturbations and outliers. These features allow multiset neurons, which are based on the coincidence similarity operation, to perform effective pattern recognition applications, including the challenging task of image segmentation. A few prototype points have been used in previous related approaches to represent each pattern to be identified, each of them being associated with respective multiset neurons.  The segmentation of the regions can then proceed by taking into account the outputs of these neurons. The present work describes multilayer multiset neuronal networks incorporating two or more layers of coincidence similarity neurons. In addition, as a means to improve performance, this work also explores the utilization of counter-prototype points, which are assigned to the image regions to be avoided. This approach is shown to allow effective segmentation of complex regions despite considering only one prototype and one counter-prototype point. As reported here, the balanced accuracy landscapes to be optimized in order to identify the weight of the neurons in subsequent layers have been found to be relatively smooth, while typically involving more than one attraction basin. The use of a simple gradient-based optimization methodology has been demonstrated to effectively train the considered neural networks with several architectures, at least for the given data type, configuration of parameters, and network architecture.
\end{abstract}

\section{Introduction}

Human intelligence is to a large extent implemented into respective \emph{neuronal networks} (e.g.~\cite{goldstein2021sensation, haykin1998neural}), which continuously treat perceptual signals while implementing learning and providing subsidies for decision-making.

Biological neuronal networks learn patterns mainly by adapting the efficiencies of the synapses of their neuronal cells in terms of a respective history or activations and learning (e.g.~\cite{haykin1998neural}). Organized along several layers, and involving about 90 billion neurons, human neuronal networks have been providing motivation for respective emulations in artificial systems, leading to the interesting area known as \emph{artificial neuronal networks} -- ANNs (e.g.~\cite{haykin1998neural,kohonen1990self}).

After several cycles of interesting developments, ANNs ultimately consolidated into the concept of \emph{deep learning} (e.g.~\cite{lecun2015deep,krizhevsky2017imagenet,pouyanfar2018survey}), in which substantial amounts of computational resources and training data are considered as the means for implementing effective and flexible pattern recognition.

Most of the ANNs have been based on a basic neuron element that incorporates an inner product at its input, implementing the combination of respectively associated weights and incoming signals, followed by a non-linear activation function such as a sigmoid or rectified linear unit (relu)~\cite{lecun2015deep}. Therefore, this amply adopted type of neuron can be understood as having a bilinear input stage implementing the inner product, which it itself directly related to the Euclidean distance.

Introduced more recently, the concept of multiset neurons, as well as their organization into respective networks \cite{costa2023multiset,costa2021multiset, benatti2023two}, have been found to allow effective and versatile pattern recognition in several types of data and applications (e.g.~\cite{costa2022on, costa2023multiset, da2022abrief}). Basically, multiset neurons correspond to computational implementations, founded on multiset theory (e.g.~\cite{costa2023multiset, da2022supervised}), of similarity measurements including the Jaccard (e.g.~\cite{Jaccard1901distribution,Jac:wiki,leydesdorff2008on,da2021further}), interiority (or overlap, e.g.~\cite{vijaymeena2016a} and coincidence indices (e.g.~\cite{da2021further,costa2022on,costa2023multiset}). 

One of the distinguishing features of a multiset neuron concerns the fact that its initial processing stage corresponds to the quantification of a respective multiset similarity between the input signals and an associated set of weights, instead of the inner product typically adopted in more traditional networks. Since the coincidence operation involves computing the minimum and maximum of pairs of values, as well as the power operation, this initial processing stage of multiset neurons is highly non-linear, enhancing the flexibility and performance of the resulting multiset neurons. This approach leads to several interesting properties while comparing two mathematic structures, including intrinsic normalization, non-dimensionality, enhanced selectivity, and sensitivity, as well as robustness to limited data perturbations and outliers~\cite{costa2022on,CostaCCompl,costa2023multiset}.

In particular, multiset neurons have allowed (e.g.~\cite{costa2023multiset, da2022supervised}) particularly accurate and effective performance of the challenging task of \emph{supervised image segmentation} (e.g.~\cite{egmont2002image, alom2018nuclei, minaee2021image,cheng2001color,pham2000current}). In this type of application, a small set of prototype points is selected from the pixels belonging to the objects to be recognized (segmented), and the properties of the neighbors of each of these prototype pixels are then taken as the \emph{weights} of respectively associated neurons. While these neurons have been mostly associated in terms of logic or operations, giving rise to a respective logical neuronal layer, it is also possible to integrate the output of prototype neurons in terms of subsequent layers of multiset neurons.

The present work provides a study of multiset neuronal networks (MNNs) incorporating two or more layers of neurons implementing the multiset coincidence similarity operator, which are henceforth referred to as \emph{multilayer multiset neuronal networks} - MMNNs. In order to allow for more flexible and effective learning, we also consider the possibility of having negative values for input as well as for weights, which requires a respective extension of the coincidence similarity index (e.g.~\cite{da2021further,costa2022on,costa2023multiset}). One immediate benefit of the latter approach is that it becomes possible to adopt not only prototype points for representing the basic patterns to be recognized but also \emph{counter-prototype points} representing patterns to be avoided. The impressive potential of this approach has been illustrated respectively to the segmentation of the leaves from an intricate image of an anthurium vase while employing only one prototype and one counter-prototype point.

Supervised pattern recognition by using MMNNs requires the adoption of a suitable and effective manner for training the weights of the neurons so as to achieve the desired operation and recognition of specific patterns, which constitutes the respective \emph{training stage}. In the present work, a gradient-based optimization methodology is considered for maximizing the accuracy of the recognition~\cite{du2019gradient,ruder2016overview}. Though all examples in the current work are respective to the important task of \emph{image segmentation}, the generalization of the described approaches, concepts, and methods to other types of data should be relatively direct.

In particular, the possibility to use only two points for training the neurons in the first neuronal layer allowed the visualization of the accuracy landscape, as well as respective trajectories delineated by the gradient ascent optimization~\cite{ruder2016overview}. At least for the considered types of images, neuronal network architectures, features, and parameter configurations, the accuracy landscapes have been found to be mostly smooth while incorporating a relatively small number of attraction basins (local extremes). The reported concepts, methods, and developments pave the way to several important subsequent studies involving larger MMNNs incorporating more neurons and layers, up to the level of deep learning approaches.

This work starts by presenting basic concepts of multisets, multiset similarity, multiset neurons, and multilayer multiset neuronal networks - MMNNs, and proceeds by illustrating the potential of the reported approaches in terms of image segmentation tasks by using prototype and counter-prototype points, as well as MMNNs trained by gradient optimization.

\section{Multisets}

In this section, some basic concepts of multiset theory adopted in the present work are reviewed in a concise and introductory manner.

Multisets (e.g.~\cite{da2021multiset,da2022multisets}) are an extension of the traditional concept of a \emph{set}, allowing for repeated elements, with the number of these repetitions constituting the respective \emph{multiplicities}. Similarly to sets, two multisets can be combined through operations, such as union, intersection, and complementation, among other possibilities.

For the sake of enhanced precision, multiset theory has been mostly approached and developed in terms of the area known as \emph{foundation of mathematics} (e.g.~\cite{blizard1989multiset, blizard1991development}), and especially modern set theories aimed at circumventing Russell's paradox involving the \emph{unrestricted comprehension principle}. Informally speaking, this inconsistency is related to recursive situations in which sets are members of themselves. Here, however, we take a simple and more direct approach to multisets (e.g.~\cite{da2021multiset, da2022multisets, da2022supervised, costa2023multiset}).

A \emph{multiset} $X$ can be represented as a set of tuples $[x_i,m_{x_i}]$, where $x_i$ corresponds to each element in $S$ while $m_i$ indicates its respective multiplicity. The set of all possible elements in the multiset $X$ is henceforth called its \emph{support} $S_X$. The total number of elements in $X$ corresponds to $N_X=|S_X|$.

As an example, consider the two following multisets:
\begin{align}
  &X = \left\{ [x_1=a, m_{x_1} = 3]; [x_2=b, m_{x_2} = 2] \right\} \\
  &Y = \left\{ [y_1=a, m_{y_1} = 1]; [y_2=b, m_{y_2} = 3]; [y_3=d, m_{y_3} = 1] \right\},
\end{align}
which can also be expressed as:

\begin{align}
  &X = \left\{ a, a, a, b, b \right\} \\
  &Y = \left\{ a, b, b, b, d \right\}.
\end{align}

Given two multisets $X$ and $Y$, their respective \emph{union} consists of the multiset $Z$ having support $S_Z = S_X \cup S_Y$ and tuples corresponding to each of the elements in $S_Z$ with multiplicity being taken as the maximum between the respective multiplicities in the original multisets $X$ and $Y$.

For instance, in the case of the above examples, we would have:
\begin{align}
  & S_{Z=X \cup Y} = \left\{a, b, d \right\} \nonumber \\
  &Z = X \cup Y = \left\{ [y_1=a, m_{x_1} = 3]; [y_2=b, m_{x_2} = 3]; [y_3=d, m_{x_3} = 1] \right\} = \nonumber \\
  & = \left\{ a, a, a, b, b, b, d \right\}.
\end{align}

The \emph{intersection} between two multisets $X$ and $Y$ corresponds to the multiset $Z$ with support $S_Z = S_X \cap S_Y$ and tuples indicating each shared element with multiplicity being taken as the minimum between the two original multiplicities in $X$ and $Y$.

In the case of the above example, it follows that:
\begin{align}
  & S_{Z=X \cap Y} = \left\{a, b \right\} \nonumber \\
  &Z = X \cap Y = \left\{ [y_1=a, m_{x_1} = 2]; [y_2=b, m_{x_2} = 2] \right\} = \nonumber \\
  & = \left\{ a, a, b, b \right\}.
\end{align}

The \emph{cardinality} of a multiset $X$, here represented as $|X|$, is henceforth taken to correspond to the sum of the multiplicities of every involved element. For instance, respectively to the above example, we would have:
\begin{align}
  &|X| = 5 \nonumber \\
  &|Y| = 5 \nonumber \\
  &|X \cup Y| = 7 \nonumber \\
  &|X \cap Y| = 4 \nonumber.
\end{align}

For generalization's sake, the multiplicity of multiset elements is henceforth extended to real values, including possible negative entries (e.g.~\cite{da2021further, CostaCCompl, costa2022on, costa2023multiset}). This allows multisets to be extended to vectors, functions, matrices, graphs, etc.

\section{Multiset Similarities}

Two traditional sets can be compared in several ways, including in terms of respective \emph{similarity indices} including Jaccard, interiority, and coincidence.

Given two non-empty sets $A$ and $B$, the \emph{Jaccard similarity index} can be respectively expressed as:
\begin{align}
   \mathcal{J}(A,B) = \frac{|A \cap B|} {|A \cup B|},
\end{align}
which is a non-dimensional commutative operation also satisfying $0 \leq \mathcal{J}(A,B) \leq 1$.

The \emph{interiority index} (also called overlap index, e.g.~\cite{vijaymeena2016a,da2021further}) would correspond to:
\begin{align}
   \mathcal{I}(A,B) = \frac{|A \cap B|} {\min \left\{ |A|, |B| \right\} },
\end{align}
which is also commutative with $0 \leq \mathcal{I}(A,B) \leq 1$.

Now, the \emph{coincidence similarity index} of the two non-empty sets $A$ and $B$ can be expressed~\cite{da2021further,costa2022on,costa2023multiset,CostaCCompl} as:
\begin{align}
   \mathcal{C}(A,B) = \left[ \mathcal{J}(A,B) \right]^D \ \mathcal{I}(A,B), \label{eq:coinc}
\end{align}
where $D \in \mathbb{R}, D> 0$ is a parameter controlling how strict the respectively implemented comparison is. The higher its value, the strict (selective and sensitive) the comparison becomes.

As with the two previous indices, the coincidence similarity is also commutative and satisfies $0 \leq \mathcal{C}(A,B) \leq 1$.

Because the three similarity indices above rely on the operations of union, intersection, and cardinality, all of which have direct multiset corresponding counterparts, it becomes possible to extend those three indices to two multisets $X$ and $Y$ with non-negative entries as follows:
\begin{align}
   &\mathcal{J}(X,Y) = \frac{\sum_{i=1}^{N} \min\left\{m_{x_i},m_{y_i} \right\}} 
   {\sum_{i=1}^{N} \max\left\{m_{x_i},m_{y_i} \right\}} \\
   &\mathcal{I}(X,Y) = \frac{\sum_{i=1}^{N} \min\left\{m_{x_i},m_{y_i} \right\}} 
   {\min \left\{ \sum_{i=1}^{|X|} m_{x_i}, \sum_{i=1}^{|Y|} m_{y_i} \right\}} \\
   &\mathcal{C}(X,Y) = \left[ \mathcal{J}(X,Y) \right]^D \ \mathcal{I}(X,Y) 
\end{align}

where $N$ is the number of elements in any of the multisets $X$ and $Y$, which are henceforth assumed to have the same number of elements.

In the case of multisets presenting eventual negative entries, the above expressions can be rewritten (e.g.~\cite{da2021further, CostaCCompl, costa2022on, costa2023multiset}) as:
\begin{align}
   &\mathcal{J}(X,Y) = \frac{\sum_{i=1}^{N} s_{i} \min\left\{|m_{x_i}|,|m_{y_i}| \right\}} 
   {\sum_{i=1}^{N} \max\left\{|m_{x_i}|,|m_{y_i}| \right\}} \\
   &\mathcal{I}(X,Y) = \frac{\sum_{i=1}^{N} \min\left\{|m_{x_i}|,|m_{y_i}| \right\}} 
   {\min \left\{ \sum_{i=1}^{|X|} |m_{x_i}|, \sum_{i=1}^{|Y|} |m_{y_i}| \right\}} \\
   &\mathcal{C}(X,Y) = \left[ \mathcal{J}(X,Y) \right]^D \ \mathcal{I}(X,Y) \label{eq:coinc_neg}
\end{align}
where:

\begin{align}
   & s_{i} = sign(m_{x_i}) \ sign(m_{y_i}).
\end{align}

\section{Multiset Neurons}

\begin{figure}
  \centering
     \includegraphics[width=0.6\textwidth]{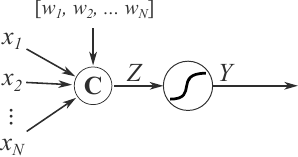}
   \caption{The main components of a multiset neuron including the coincidence similarity operator implementing the comparison between the input $\vec{x} = \left(x_1, x_2, \ldots, x_N \right)$ and the respective weights $\vec{w} = \left( w_1, w_2, \ldots, w_N \right)$, yielding the respective scalar value $Z$ which then undergoes a possibly non-linear activation function (e.g.~a sigmoid function, in the case of this particular example) generating the neuronal output $Y$ quantifying the overall similarity between the input and weight vectors.}\label{fig:MultiSets}
\end{figure}

Basically, a multiset neuron corresponds to the implementation of a coincidence multiset similarity index at its initial processing stage, eventually followed by a non-linear activation function. Figure~\ref{fig:MultiSets} illustrates a coincidence similarity multiset neuron with a sigmoid activation function. Other types of similarities (e.g.~Jaccard, coincidence) and activation functions (e.g.~linear, sigmoid, relu) can be readily adopted.

The neuronal input consists of $N$ input real-valued signals $x_i$, $i=1, 2, \ldots, N$, yielding a respective input vector $\vec{x}$. This input is compared with a respective weight vector $\vec{w}$, also with $N$ elements $w_i$, in terms of the coincidence similarity index (Eq.~\ref{eq:coinc}), yielding a respective scalar value $y$, with $-1 \leq Y \leq1$, which then undergoes a linear or non-linear activation function, resulting in the neuronal output $Z$.

In this work, we consider a linear activation function as well as the following sigmoid:
\begin{equation}
    Y = \frac{1}{1-e^{-a(Z+b)}},
\end{equation}
where $a$ and $b$ are free parameters.

The output of the multiset neuron above indicates the similarity between the input pattern and the weight vector. A straightforward approach for training is to take the weight as a prototype point of the pattern to be recognized. Cases involving more than one prototype point per pattern can be addressed by associating a neuron with each prototype as described above. Other training approaches, such as gradient-based optimization, can also be used to train individual multiset neurons.

\section{Multiset Neuronal Networks -- MNNs}\label{sec:Prototype}

Traditional models of individual neuronal cells, such as the McCulloch and Pitts (e.g.~\cite{lettvin1959frog}), incorporate two main stages: (i) a weighted sum of the inputs, typically implemented in terms of the inner product between the input and synaptic weights vector; and (ii) an activation function, frequently non-linear. Given that the inner product implements a comparison between the angles of two vectors, the initial stage can be understood as corresponding to a comparison between the input pattern and the respective weight vector.

Similarity indices, such as Jaccard, provide a quantification of the similarity between two sets. By substituting set operations with respective multiset counterparts (e.g.~\cite{da2022multisets}), it is possible to obtain respective similarity indices which can be applied to real-valued vectors~\cite{da2021further, CostaCCompl, costa2022on, costa2023multiset}.

The fact that similarity indices implement a comparison between two non-zero vectors motivated the concept of \emph{multiset neurons} and respective networks (e.g.~\cite{costa2023multiset, benatti2023two}), in which the traditionally adopted inner product (also related to similarity between vectors) in the first stage is replaced by a similarity index. Two types of multiset neurons have received particular attention, being respectively based on the Jaccard (e.g.~\cite{Jaccard1901distribution, leydesdorff2008on, Jac:wiki, da2021further}) and coincidence similarity indices (e.g.~\cite{costa2022on,da2021further}). Artificial neuronal networks based on multiset neurons have been called \emph{multiset neuronal networks} -- MNNs.

The potential of MNNs stems from the interesting properties of the Jaccard and coincidence similarity indices, including intrinsic normalization, relatively high selectivity, and sensitivity, as well as tolerance to data perturbations and the presence of outliers (e.g.~\cite{costa2022on, costa2023multiset}). In particular, the \emph{coincidence similarity index} has been introduced as a means to complement the Jaccard index so that the relative interiority between the two multisets can also be taken into account~\cite{da2021further}. As a consequence, the coincidence similarity index performs an even more strict comparison between two non-zero vectors.

It is also interesting to observe that the Jaccard and coincidence similarity indices are intrinsically non-linear, while the inner product is a bi-linear operation. This property contributes to the enhanced selectivity and flexibility of the comparisons implemented by the MNNs, as well as to other interesting features.

Effective performance of multiset neurons has been observed~\cite{costa2023multiset, da2022supervised, benatti2023two}  respectively to the particularly challenging task of \emph{image segmentation} (e.g.~\cite{gonzales1987digital, costa2000shape}), in which portions of interest (e.g.~objects against a background, or parts of objects) in a gray-level or color image are to be recognized. Previous applications of MNNs to supervised image segmentation~\cite{costa2023multiset, da2022supervised, benatti2023two} have been based on associating prototype (or sample) points, indicating pixels that are characteristic of the regions to be segmented, to respective multiset neurons with weights corresponding to a set of features characterizing the respective prototype point. Features of particular interest which have been adopted so far~\cite{costa2023multiset, da2022supervised, benatti2023two} include the gray-level or color values within a neighborhood of each pixel, as well as the distance from the central pixel and those in the respective neighborhood.

Prototype points have been set interactively by human supervision, which constitutes the MNN training stage. Basically, a relatively small number of prototype points are interactively chosen (e.g.~through mouse clicking) so as to represent particularly characteristic points associated with the regions to be segmented.

In order to reduce the time required for the training stage, it is possible to consider a sub-sampled version of the image and respective segmentations, by a factor of 10. This approach has been adopted in the present work.

The output of these multiset neurons is then combined by using logical operations, such as the \emph{or}, so that a single overall scalar value can be obtained indicating the possible presence of the sought region (e.g.~object) in the presented image. These indications are respective to each of the image pixels (or a subset of special interest), which are thus sequentially scanned as input into the respective MNN, therefore yielding a new respective image identifying the recognized regions in terms of 1s and 0s.

To evaluate the accuracy of binary image segmentation, we use the Balanced Accuracy ($BA$) metric~\cite{garcia2009index}. The Balanced Accuracy is defined as the average of two metrics, the sensitivity, and the specificity: 
\begin{equation}
   BA =\frac{1}{2}\frac{TP}{(TP+FN)} + \frac{1}{2}\frac{TN}{(TN+FP)},
\end{equation}

where TP = \emph{True Positive}; FP = \emph{False Positive}; TN = \emph{True Negative}; FN = \emph{False Negative}.

\section{Counter-Prototype Points}\label{sec:Counter-Prototype}

While the previous developments concerning the use of multiset neurons and MNNs to pattern recognition have mostly adopted input and weights with strictly non-negative or positive entries, more flexible and powerful recognition can be potentially obtained by considering also negative entries, so that real-valued inputs and weights are allowed. The main interesting feature allowed by this extension consists in the possibility of using not only prototype points but also \emph{counter-prototype points} indicating patterns to be avoided. These possibilities are addressed in the present section as a preliminary step before proceeding to MMNNS.

The extension of multiset neurons and respective networks to real-valued inputs and weights can be readily achieved by considering the extended coincidence similarity index in Equation~\ref{eq:coinc_neg}.

The counter-prototype points are trained in the same way as the prototype points. More specifically, in the case of image segmentation, the gray-level or color properties of the pixels within a circular region with radius $r$ centered at each of the reference pixels are vectorized into a respective \emph{feature vector}, which can also incorporate sorting the elements.

\begin{figure}
  \centering
     \subfloat[]{\includegraphics[width=0.4\textwidth]{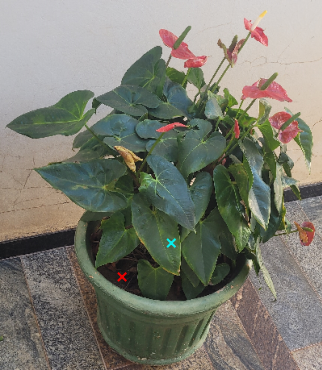}}
     \hspace{0.7cm}
     \subfloat[]{\fbox{\includegraphics[width=0.384\textwidth]{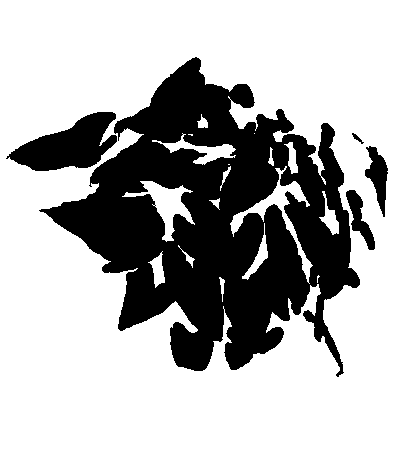}}}
     
     \vspace{0.9cm}
     
     \subfloat[]{\fbox{\includegraphics[width=0.384\textwidth]{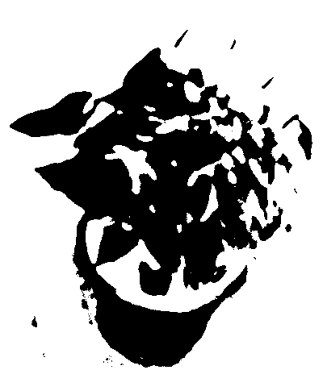}}}
     \hspace{0.7cm}
     \subfloat[]{\includegraphics[width=0.4\textwidth]{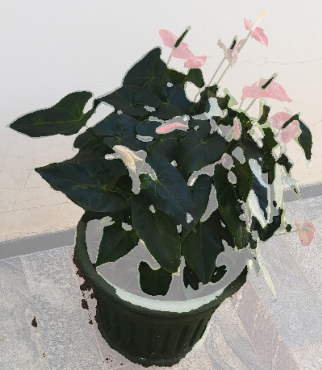}}
     
   \caption{Example of segmentation of the leaves of an anthurium plant in a vase while considering only two points: one prototype (cyan) and one counter-prototype (red). The original HSV image is shown in (a), with the respective human-defined gold standard being shown in (b). The segmentation results considering the two points and parameter configuration $D = 3$, $r = 4$, $a = 2000$, $b = 0.0$, and $T = 0.02$ is shown in (c), being characterized by a balanced accuracy of 86.47 \%. Image (d) presents the segmentation in (c) superimposed onto the original image in (a). Except for the incorporation of the vase, which has the same color properties as the leaves, a markedly good segmentation result has been obtained in spite of the color and intensity variations and intricate shape of the leaves. Observe that the undetected leaves basically correspond to intense reflections that are nearly white. Even better results can be obtained by using more prototype and counter-prototype points.}
  \label{fig:vase}
\end{figure}

As an application example of the above approach to a real-world image, Figure~\ref{fig:vase} presents the segmentation (b) of the leaves in a vase in the image shown in (a) by considering just one prototype and one counter-prototype point, shown in cyan and red, respectively. This task is particularly challenging because of the irregular border of the region of interest, as well as the intense variation of its visual properties (color, gray level, texture, reflections, etc.).

An impressively accurate result has been obtained despite variations in the color and intensity of the leaves, as well as their intricate borders, reflections, shadows, and proximity to regions with similar properties (e.g.~earth and floor). A good portion of the vase resulted incorporated into the segmented region because of the great overlap of its color properties with those of the leaves.  Additional features about the image (e.g.~curvature, textures, relative position, etc.) would be needed to be taken into account in case the vase is to be avoided. The small portions of the leaves which have not been identified can be found to correspond to reflections. Observe that these regions can only be properly identified by incorporating high-level intelligence about the typical shapes of the anthurium plant, as well as illumination models.

\begin{figure*}
  \centering
     \includegraphics[width=0.95\textwidth]{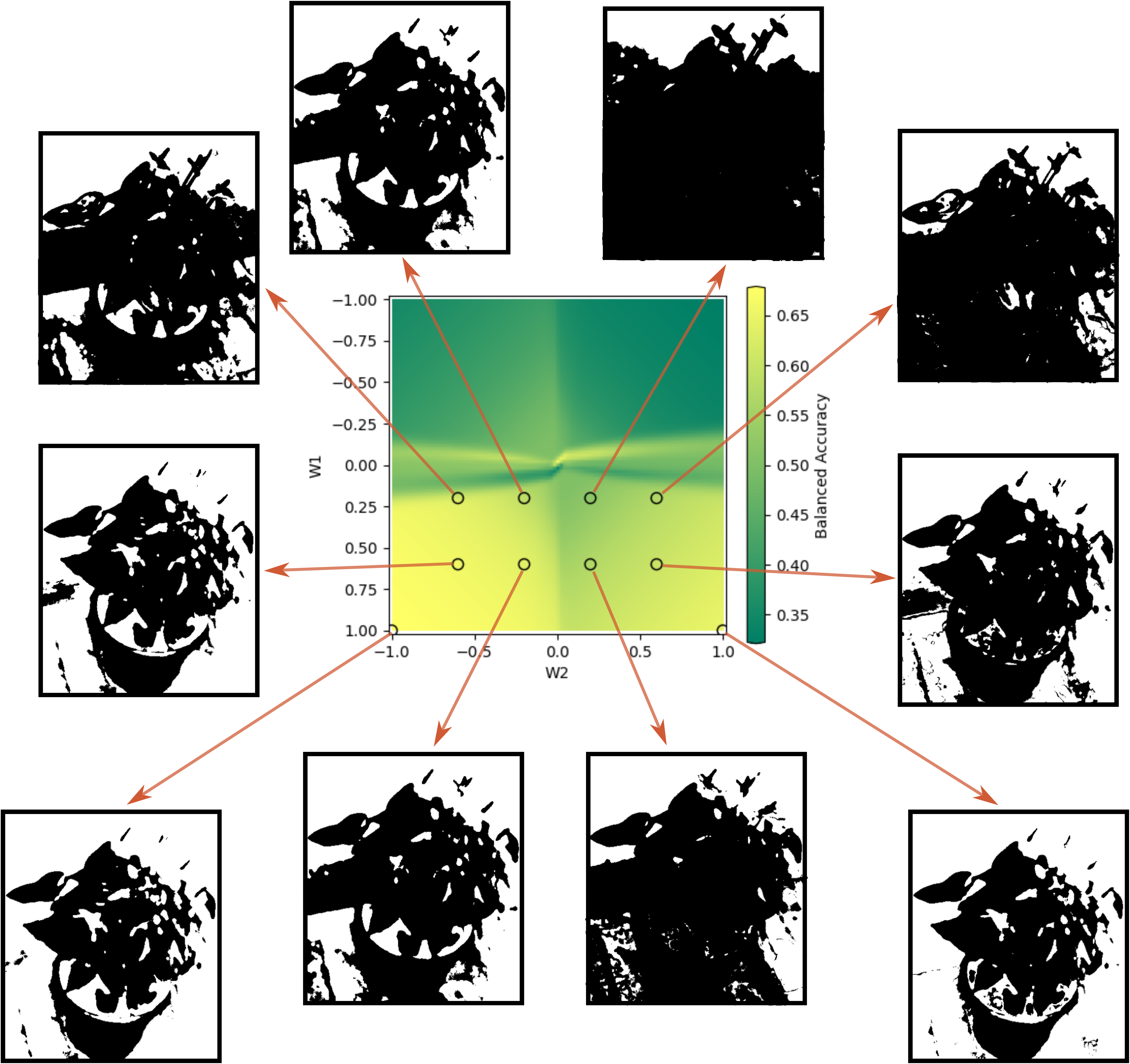}
   \caption{The balanced accuracy landscape obtained for the segmentation of the leaves in the anthurium image in Fig.~\ref{fig:vase} and a set of 10 segmentations obtained for respective weight configurations $(w_1,w_2)$. The two best results have been obtained for $(w_1=1,w_2=-1)$ and $(w_1=1,w_2=1)$.}\label{fig:vase_grad}
\end{figure*}

Figure~\ref{fig:vase_grad} depicts at its center the balanced accuracy landscape, as well as 10 segmentation results obtained from respective weight configurations as indicated by the arrows. The upper half of the accuracy landscape has not been considered as it yields only blank images. It can be observed that varying levels of accuracy can be obtained for distinct weight configurations, with the best result being observed for the specific configuration $(w_1=1,w_2=-1)$.

\begin{figure}
  \centering
     \subfloat[]{\includegraphics[width=0.4\textwidth]{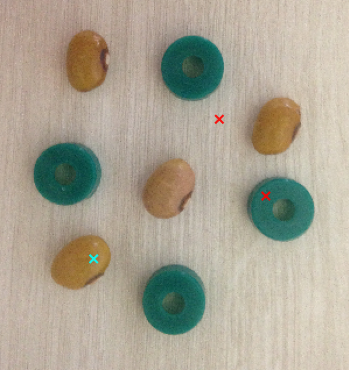}}
     \hspace{0.7cm}
     \subfloat[]{\fbox{\includegraphics[width=0.384\textwidth]{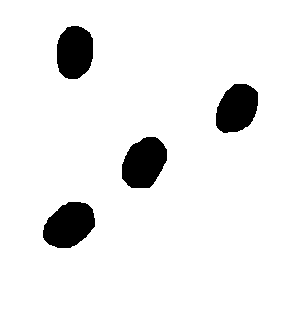}}}
     
     \vspace{0.9cm}
     
     \subfloat[]{\fbox{\includegraphics[width=0.384\textwidth]{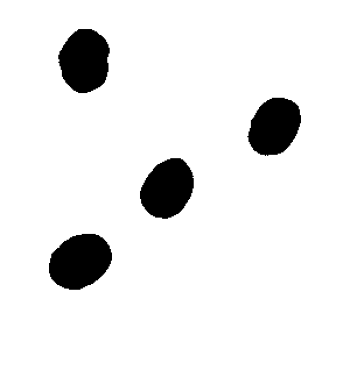}}}
     \hspace{0.7cm}
     \subfloat[]{\includegraphics[width=0.4\textwidth]{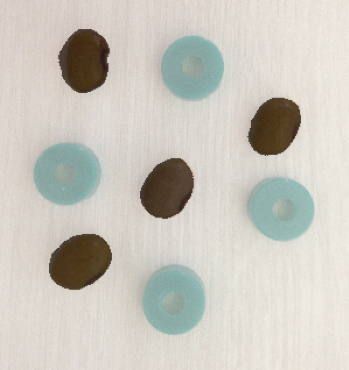}}
     
   \caption{Segmentation of the beans from the image shown in (a) while considering only one prototype (cyan) and two counter-prototype (red) points. The respective gold standard image, obtained by human inspection is shown in (b), and the segmentation results obtained for the parameter configuration $D = 5$, $r = 3$, $a = 5,000$, $b = 0$, and $T = 0.5$ is depicted in (c), which yielded an accuracy as high as $BA = 98.79$ \%. The image in (d) illustrates the superimposition of the segmented beans onto the original image.}
  \label{fig:img}
\end{figure}

\begin{figure}
  \centering
     \subfloat[]{\includegraphics[width=0.4\textwidth]{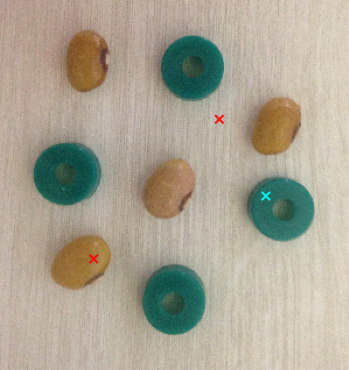}}
     \hspace{0.7cm}
     \subfloat[]{\fbox{\includegraphics[width=0.384\textwidth]{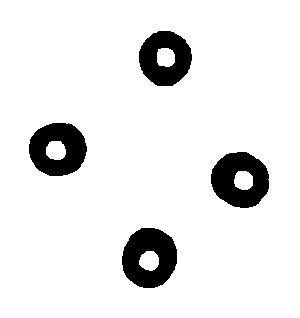}}}
     
     \vspace{0.9cm}
     
     \subfloat[]{\fbox{\includegraphics[width=0.384\textwidth]{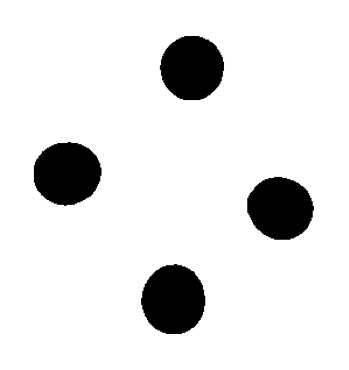}}}
     \hspace{0.7cm}
     \subfloat[]{\includegraphics[width=0.4\textwidth]{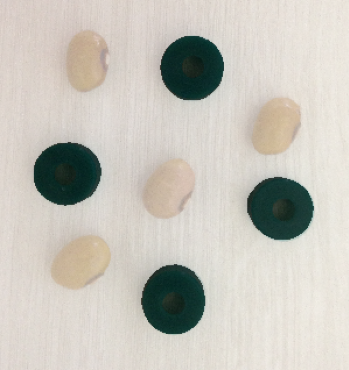}}
     
   \caption{Segmentation of the green washers from the image shown in (a) while considering only one prototype (cyan) and two counter-prototype (red) points. The respective gold standard, obtained by human inspection is shown in (b), and the segmentation results obtained for the parameter configuration $D = 5$, $r = 3$, $a = 5,000$, $b = 0$, and $T=0.5$ is depicted in (c), which yielded an accuracy as high as $BA = 97.57$ \%. The holes through the washers could not be detected because of the intense respective shadow and consideration of counter-prototype points respective only to the bean and background. The image in (d) illustrates the superimposition of the segmented washers onto the original image.}
  \label{fig:img2}
\end{figure}

Figure~\ref{fig:img} presents additional illustrations of the application of MNNs to the identification of beans, placed against a background and in presence of other objects (wahsers), while considering one prototype (cyan) and two counter-prototype (red) points. The recognition of the washers is depicted in Figure~\ref{fig:img2}.

\section{Gradient Optimization in Two-Layer MNNs} \label{sec:Two-Layer}

Though the training of individual multiset neurons can be readily implemented by using respective prototype and counter-prototype points (see Sections~\ref{sec:Prototype} and~\ref{sec:Counter-Prototype}), more effective and flexible results can be obtained by employing several neurons. However, combining two or more individual multiset neurons requires additional neuronal layers and the adoption of a specific training methodology. Two of the main possibilities for implementing additional layers include: (i) combining logically thresholded versions of the outputs of each individual multiset neuron; and (ii) combining versions of the individual multiset neurons by using a second layer of multiset neurons with suitably identified respective weights and a non-linear output stage.

While alternative (i) has been used previously~\cite{costa2023multiset, da2021multiset, benatti2023two}, and the adoption of non-linear output stages has been described in~\cite{costa2023multiset}, alternative (ii) is further explored in the present work.

\begin{figure}
  \centering
     \includegraphics[width=0.95\textwidth]{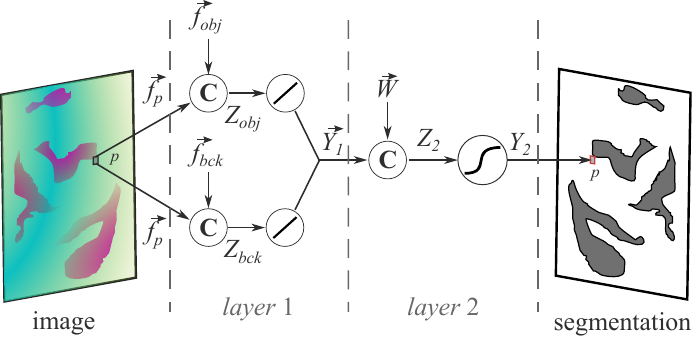}
   \caption{A simple MNN architecture composed of two layers. The first layer corresponds to two coincidence multiset neurons (with linear activations) trained with respective prototypes representing the object to be segmented and the background. These two neurons compare the feature vectors $\vec{f}_p$ of each pixel in the image, which is followed by a linear activation function. The output from layer 1 is organized into the vector $\vec{Y}_1$, which is then input into the second layer, containing a single coincidence multiset neuron (with sigmoid activation) that compares this input vector with the respective weight vector $W$ and outputs a respective scalar value $Z_2$ which, after undergoing a non-linear activation function, yields the overall output $Y_2$. High values of $Y_2$ observed for each pixel $p$ indicate that it belongs to the segmented region.}\label{fig:2layers}
\end{figure}

Figure~\ref{fig:2layers} illustrates a two-layer MNN involving two multiset neurons at the first neuronal layer with linear outputs, as well as another multiset neuron with non-linear output constituting the second neuronal layer.

It is henceforth assumed that the two neurons in the first layer have been trained by using weights corresponding to the features of respective prototypes of the two types of patterns to be recognized, $\vec{f}_{obj}$ and $\vec{f}_{bck}$, which constitutes the \emph{training stage}.

The features of each of the pixels $p$ in the image are then scanned (in any order) as input to the two multiset neurons in the first layer, yielding the respective coincidence values $Z_{obj}$ and $Z_{bck}$, each of them a real-valued scalar. Because both these neurons have linear activation functions, the outputs of the first layer can be represented in terms of the two-entries vector $\vec{Y}_1$, which is then input to the second neuronal layer, where it is compared, via the coincidence index, with the weight vector $\vec{W}$. The respective scalar output subsequently undergoes a non-linear activation function, yielding the overall output $Y_2$.

It is understood that $Y_2$ should be high whenever the input contains entries similar to the two sought patterns, which requires the weights $\vec{W}$ of the neuron in the second layer to be defined so as to achieve this type of operation. In the present work, this will be done by implementing a gradient-based optimization approach.

The optimization of weights in artificial neuronal networks constitutes a potentially challenging task because of the presence of \emph{local extremes} which can substantially hinder the identification of the absolute extreme associated with the optimal configuration of a given network respectively to a specific pattern recognition application.
Though several methods have been developed in order to address this important limiting issue, including simulated annealing~\cite{press2007numerical}, ultimately there is no way to ensure that the absolute optimal point will be found in a finite period of time.

In this study, the scalar field being optimized refers to the \emph{balanced accuracy} (e.g.~\cite{garcia2009index}) obtained in terms of the weights from training images and their respective gold standards. Both images have dimensions of $N_x \times N_y$. To be specific, we add the output of the second layer of the MNN within the region of the gold standard image that corresponds to the object being sought, and then subtract the total neuronal output obtained for the background region of the same gold standard image. Therefore, the resulting value of this subtraction, identified as $A$, provides a quantification of the overall accuracy of image recognition for each distinct configuration of weights $\vec{W}$ in the second neuronal layer.

Greater values of $A$ indicate higher balanced accuracy in the recognition process. Consequently, we implement a gradient ascent methodology (e.g.~\cite{press2007numerical}), which is applied to a set of $P$ distinct initial points. Thus, $P$ trajectories are obtained within the weight space at discrete time steps. Each trajectory is allowed to proceed until a change in $A$ becomes smaller than a threshold value of $T_A$. We select the solution with the highest value of $A$ as the trained weights.

In this work, we start from a set of 10 randomly chosen initial guesses of the parameters, from which a gradient ascent optimization is performed along 30 consecutive steps (with resolution $\delta w=0.01$), respectively to the maximization of the balanced accuracy. The learning rate, which is the size of the steps in each iteration, was fixed at $\Delta A = 0.05$.

\begin{figure*}
  \centering
     \subfloat[]{\includegraphics[width=0.4\textwidth]{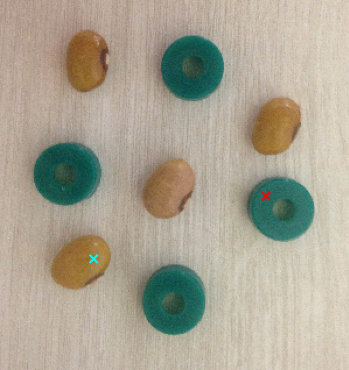}} \hspace{1cm}
     \subfloat[]{\includegraphics[width=0.4\textwidth]{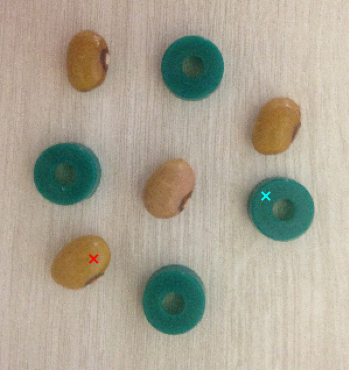}}
   \caption{Real-world HSV image containing four beans and four washers grains, used in this work to illustrate the gradient ascent approach to training a two-layer coincidence MNN. The prototype (cyan) and counter-prototype (red) points adopted for the identification of the beans and washers are respectively shown in (a) and (b).} \label{fig:img_TwoPoints}
\end{figure*}

\begin{figure}
  \centering
     \includegraphics[width=0.8\textwidth]{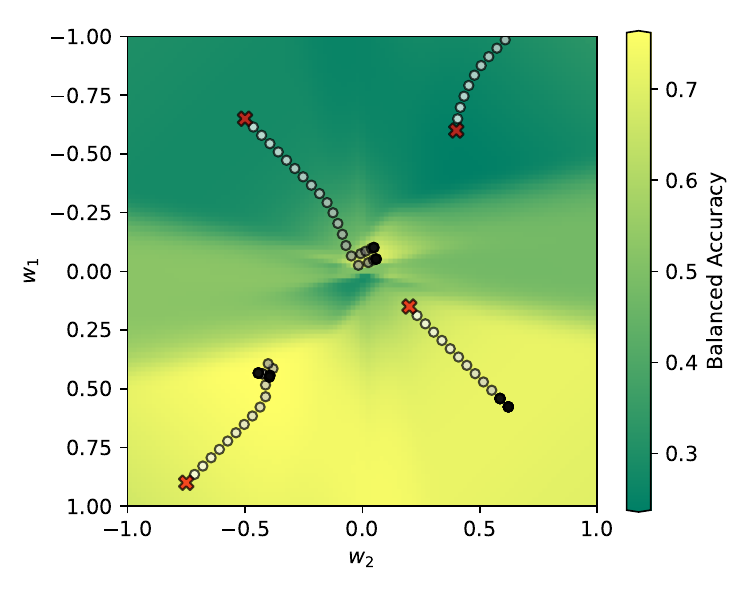}
   \caption{The balanced accuracy surface values (represented as a heatmap) obtained for the recognition of the beans in Figure~\ref{fig:img_TwoPoints}(a) by using two coincidence neurons respectively to the prototype and counter-prototype points shown in blue and red. Also shown are four trajectories defined by gradient ascents starting at respective initial points (in red).}\label{fig:img1_grad}
\end{figure}

\begin{figure}
  \centering
     \includegraphics[width=0.8\textwidth]{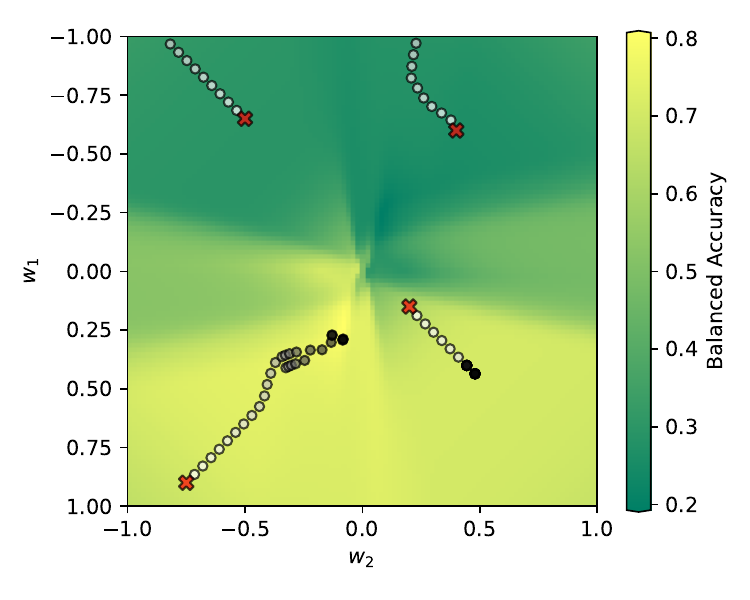}
   \caption{The balanced accuracy surface values (represented as a heatmap) obtained for the recognition of the washers in Figure~\ref{fig:img_TwoPoints}(b) by using two coincidence neurons respectively to the prototype and counter-prototype points shown in blue and red. Also shown are four trajectories defined by gradient ascents starting at respective initial points (in red). }\label{fig:img2_grad}
\end{figure}

Interestingly, the network architecture illustrated in Figure~\ref{fig:2layers} incorporates a single 2D weight vector $\vec{W} = [w_1,w_2]$, which allows the complete visualization of the respective $A$ surface. Figures~\ref{fig:img1_grad} and~\ref{fig:img2_grad} illustrate this surface respectively to the recognition of the beans and corn grains in Figure~\ref{fig:img_TwoPoints}. A level set visualization of the balanced accuracy landscape is also illustrated in Figure~\ref{fig:img1_grad}.

\begin{figure}
  \centering
     \includegraphics[width=0.8\textwidth]{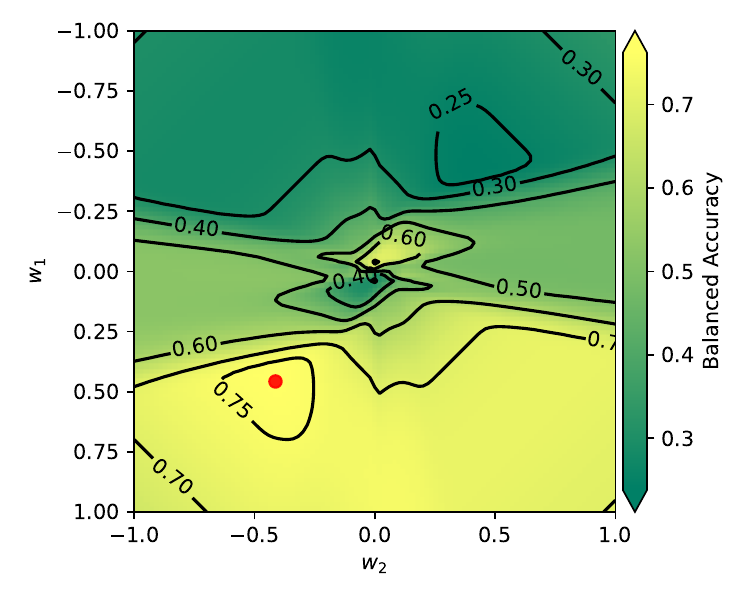}
   \caption{Visualizing the balanced accuracy in Figure~\ref{fig:img1_grad} in terms of respective level sets allows a more complete understanding of the respective geometry and attraction basins. The point leading to the absolute maximum accuracy is shown as a red dot.}\label{fig:curve_A.}
\end{figure}

Also shown in Figures ~\ref{fig:img1_grad} and~\ref{fig:img2_grad} are four gradient ascent trajectories obtained respectively to four starting points (shown in red). The two obtained balanced accuracy surfaces appear to be similar, but they are actually distinct. This is more discernible from the different trajectories each method produces from the same four initial points.

Each of the trajectories shown in Figures~\ref{fig:img1_grad} and ~\ref{fig:img2_grad} moved toward different places (the initial points of each trajectory are marked in red). Nevertheless, the overall peak of balanced accuracy was properly identified by the trajectory starting at point $[0.90,-0.75]$.

\begin{figure}
  \centering
     \includegraphics[width=0.85\textwidth]{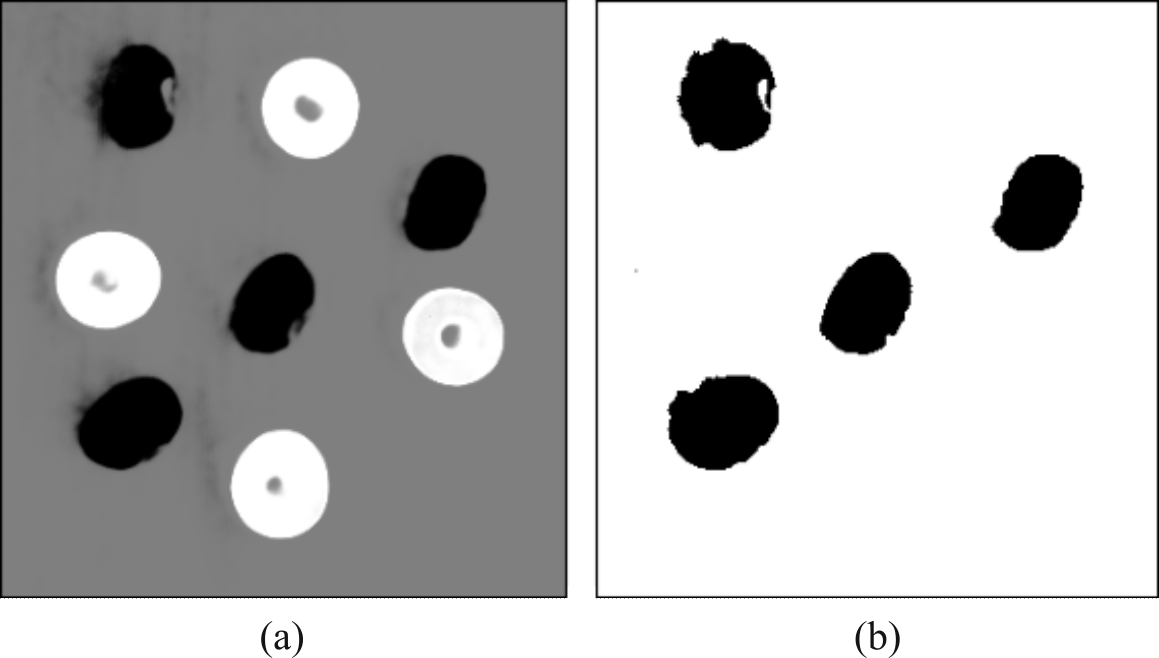}
   \caption{(a): Result of segmentation of the \emph{beans} in Fig.~\ref{fig:img_TwoPoints}(a) obtained by using the MNN in Fig.~\ref{fig:img1_grad} with parameters optimized by gradient ascent as $\vec{W} = [0.457, -0.413]$. (b): The thresholded version of the image in (a) for $T = 0.4$. The achieved balanced accuracy was 98.51 \%.}\label{fig:seg_beans}
\end{figure}

Figure~\ref{fig:seg_beans}(a) depicts the signals $Y_2$ obtained for the beans in Figure~\ref{fig:img_TwoPoints}(a). A respective thresholded version is shown in Figure~\ref{fig:seg_beans}(b).

\begin{figure}
  \centering
     \includegraphics[width=0.85\textwidth]{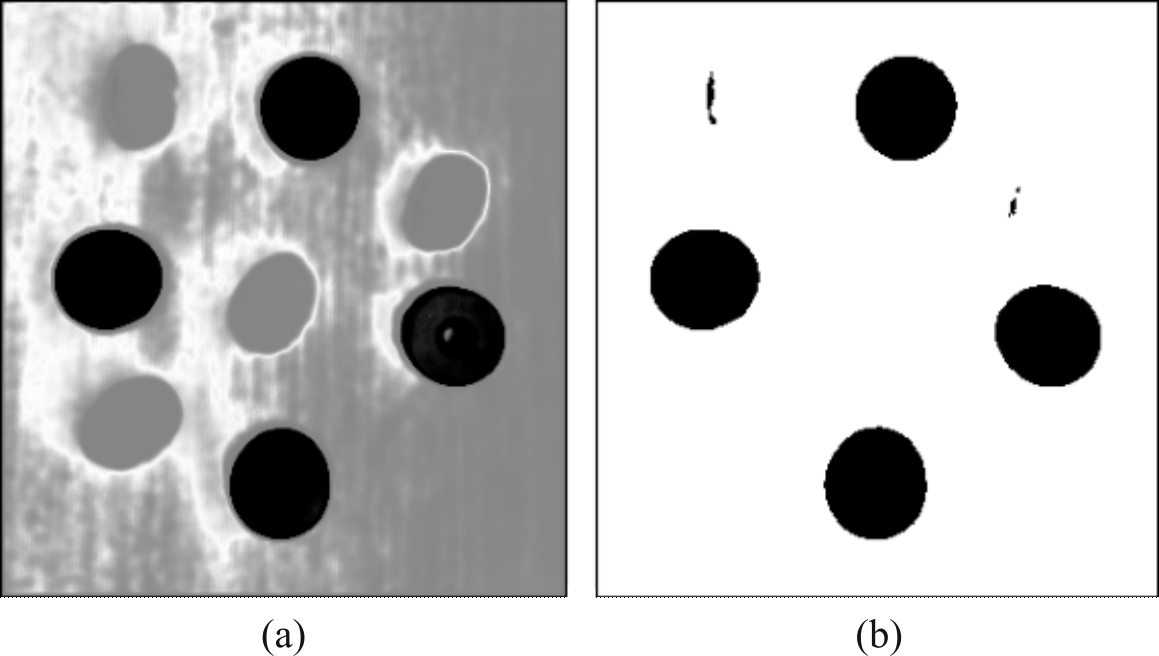}
   \caption{(a): Result of segmentation of the \emph{washers} in Fig.~\ref{fig:img_TwoPoints}(b) obtained by using the MNN in Fig.~\ref{fig:img2_grad} with parameters optimized by gradient ascent as $\vec{W} = [0.275, -0.087]$. (b): The thresholded version of the image in (a) for $T = 0.49$. The achieved balanced accuracy was 97.35 \%.}\label{fig:seg_washers}
\end{figure}

Figures~\ref{fig:seg_washers}(a) and (b) present the results obtained analogously for the washers.

Considering that only two training points have been considered for each of the two types of grains, the segmentation results can be considered to be markedly good.

\begin{figure}
  \centering
     \subfloat[]{\includegraphics[width=0.7\textwidth]{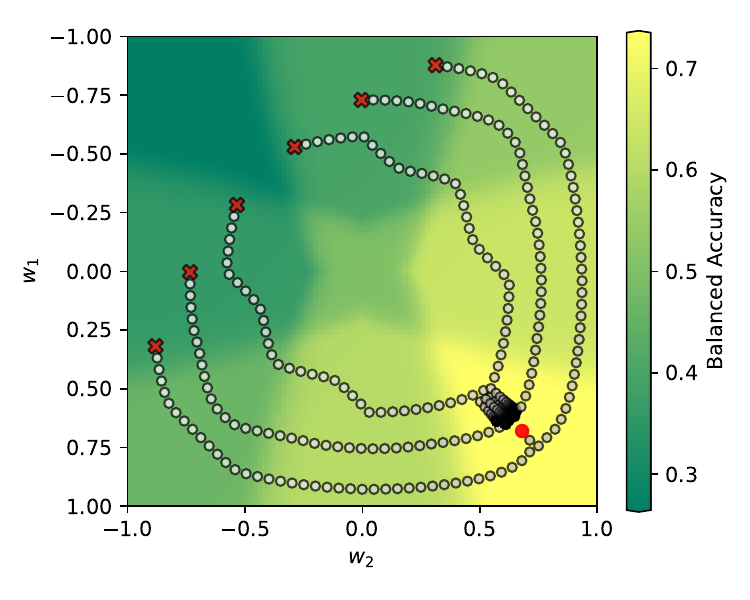}}
     \\
     \subfloat[]{\includegraphics[width=0.7\textwidth]{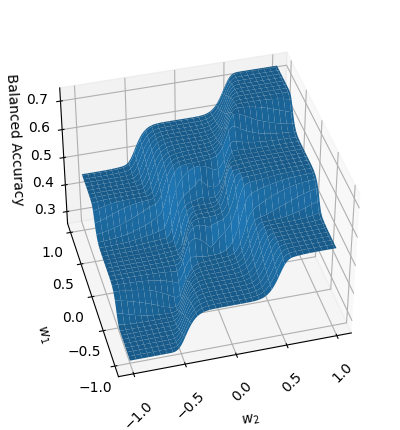}}
   \caption{(a): Examples of gradient-ascent trajectories obtained for the balanced accuracy landscape defined by the joint identification of beans and washers. (b): Three-dimensional visualization of the above-mentioned landscape reveals the presence of several adjacent plateaus.}\label{fig:img_grad2}
\end{figure}

Now, to further illustrate the application of gradient-ascent optimization for training multiset neurons, we consider the combination, in terms of an additional multiset neuron, of thresholded versions of previously obtained segmentation of the beans and washers. Figure~\ref{fig:img_grad2}(a) presents the respective landscape of balanced accuracy, as well as examples of six trajectories delineated by respective gradient-ascent optimization starting at the red `$\times$'. The maximum accuracy value was obtained for coordinates (0.68,0.68).

A better idea of the involved accuracy landscape can be obtained from the 3D visualization shown in Figure~\ref{fig:img_grad2}(b), which reveals that this landscape involves several adjacent plateaus, one of them corresponding to an almost plane region where the maximum accuracy is obtained. Interestingly, all considered trajectories converged to this maximum region.

\begin{figure}
  \centering
     \subfloat[]{\fbox{\includegraphics[width=0.384\textwidth]{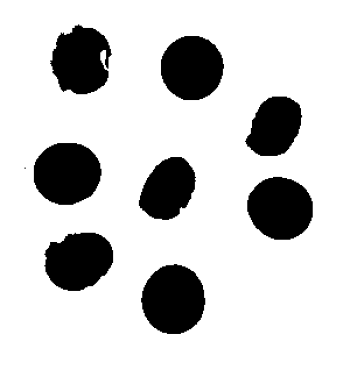}}}
     \hspace{0.7cm}
     \subfloat[]{\includegraphics[width=0.4\textwidth]{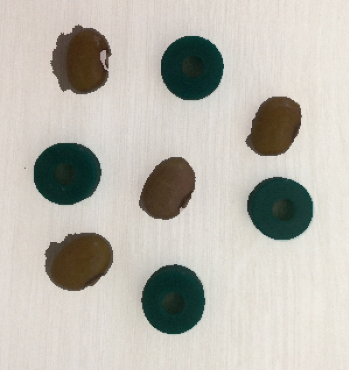}}
   \caption{Result of the joint segmentation of beans and washers (a) obtained by an additional multiset neuron acting on the thresholded respective segmented images (Figs.~\ref{fig:seg_beans}(b) and \ref{fig:seg_washers}(b)), as well as respective superimposition onto the original image (b).}
  \label{fig:img3}
\end{figure}

The result of the joint segmentation of the beans and washers is illustrated in Figure~\ref{fig:img3}, which indicates good overall performance, with a balanced accuracy of 73.42 \%

\section{Gradient Descent in Three-Layer MNNs}\label{sec:Three-Layer}

\begin{figure}
  \centering
     \includegraphics[width=0.95\textwidth]{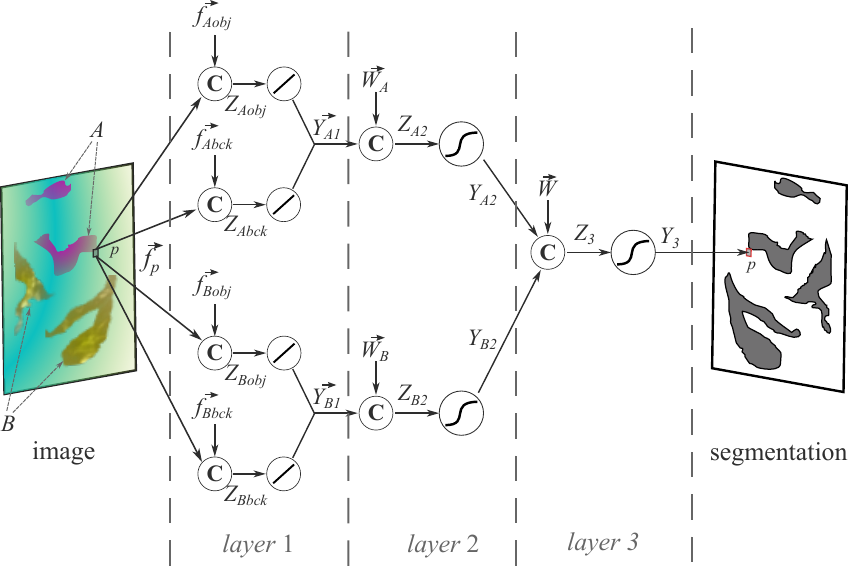}
   \caption{Architecture of a 3-layer MNN for identifying (segmenting) the two types of patterns (A and B) from the original image. The weights in the first neuronal layer are trained from respective prototypes and counter-prototypes, but the weights in the second and third layers are trained jointly by using gradient-ascent optimization.}\label{fig:3layers}
\end{figure}

In order to further study the use of gradient-ascent optimization in MMNNs, we now proceed to a 3-layer architecture as illustrated in Figure~\ref{fig:3layers}. At the first layer, a pair of neurons are assigned respectively to prototype and counter-prototypes obtained from two regions A and B (shown in magenta and yellow). The two neurons in the second layer implement the respective segmentation of the two regions, which are then combined by the neuron in the third layer.

\begin{figure}
  \centering
     \subfloat[]{\includegraphics[width=0.4\textwidth]{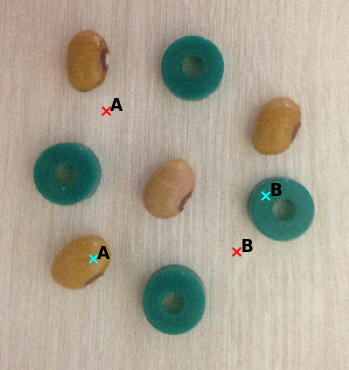}}
     \hspace{0.7cm}
     \subfloat[]{\fbox{\includegraphics[width=0.384\textwidth]{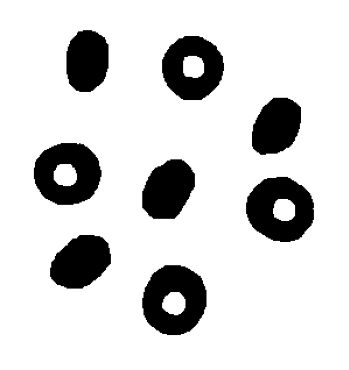}}}
     
     \vspace{0.9cm}
     
     \subfloat[]{\fbox{\includegraphics[width=0.384\textwidth]{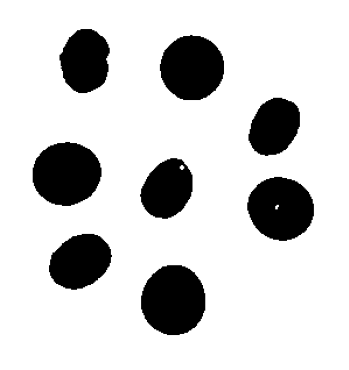}}}
     \hspace{0.7cm}
     \subfloat[]{\includegraphics[width=0.4\textwidth]{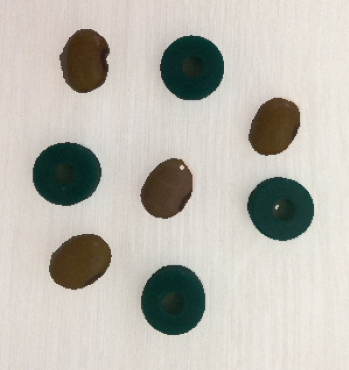}}
     
   \caption{Results of the joint segmentation of the beans and washers obtained by gradient ascent simultaneous optimization of the weight of the neurons in the second and third layer of the architecture in Fig~\ref{fig:3layers}. The obtained parameter configuration was $D=5$, $r=3$, $a=5,000$, $b=0$, and $\vec{W}_A = [-0.283, 0.106]$, $\vec{W}_B = [ 0.312, -0.320]$, $\vec{W}=[0.3823, -0.802]$ resulting in a balanced accuracy of 96.80 \%.}
  \label{fig:img_3layars}
\end{figure}

Gradient-ascent methodology has been employed to train the weights of all three neurons in the second and third layers simultaneously. The respectively obtained results, shown in Figure~\ref{fig:img_3layars}, can be verified to be good and even more accurate (balanced accuracy of 96.80 \%) than the joint segmentation obtained from the thresholded versions of the separated segmentation of the two types of objects as described in Section~\ref{sec:Two-Layer}.

\section{Generic MMNN Architecture}

\begin{figure}
  \centering
     \includegraphics[width=0.9\textwidth]{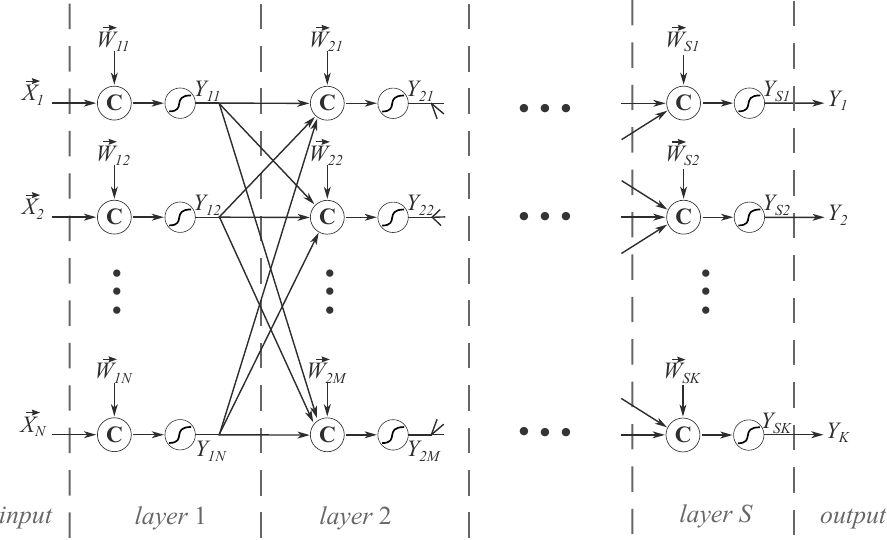}
   \caption{Generic feed-forward architecture of a multilayer multiset neuronal network, incorporating $S$ layers and a variable number of neurons per layer. Though the activation functions shown in the diagram are sigmoids, other types of functions can also be employed. The illustrated architecture receives $N$ feature vectors as input, resulting in $K$ scalar outputs. The training of the network consists in determining the involved weights by using some optimization method, such as gradient ascent.}\label{fig:NeuralNe}
\end{figure}

Several progressively more complete MMNNs architectures have been discussed and illustrated along the previous sections, proceeding from two to three layers. The most general feed-forward MMNN architecture is addressed in the present section, which is illustrated in Figure~\ref{fig:NeuralNe}.

This generic architecture involves $S$ layers, with a varying number of neurons within each of them. A total of $N$ feature vectors are taken as input, generating $K$ respective scalar values as output. The weights are trained by using some optimization approach, such as the gradient ascent methodology adopted in the present work. Several possible types of activation functions can be adopted, including the sigmoid function illustrated in the figure, as well as the relu.  Architectures incorporating several types of activation functions are also possible. Interestingly, the adoption of linear activation functions is justified because, unlike the inner product adopted in more traditional neurons, the similarity comparison constituting the first processing stage of multiset neurons is intrinsically non-linear.

In addition to incorporating a generic number of layers and neurons, the generic architecture differs from those addressed in the previous sections also by using an optimization methodology to determine any of the involved weights, including those at the first layer.

\begin{figure}
  \centering
     \subfloat[]{\includegraphics[width=0.95\textwidth]{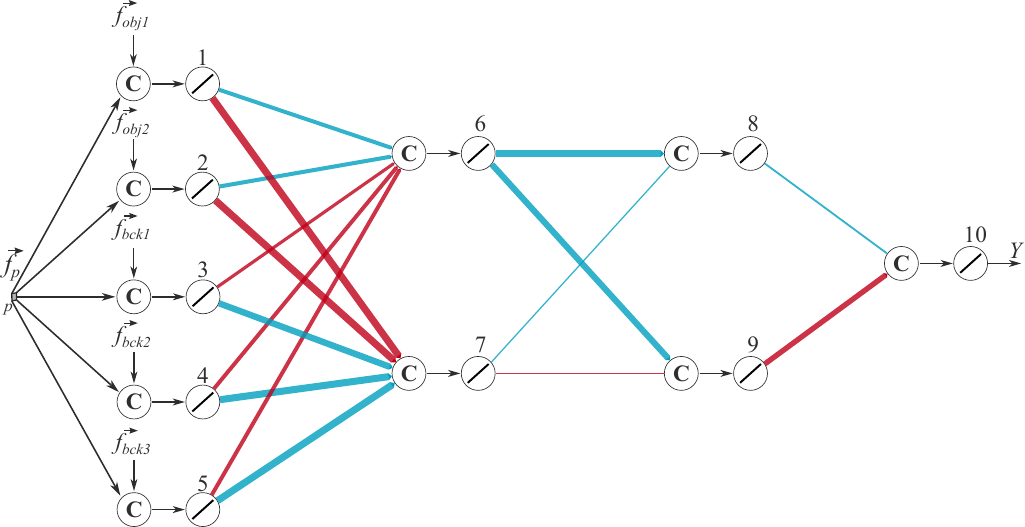}}\\ \vspace{.5cm}
     \subfloat[]{\includegraphics[width=0.95\textwidth]{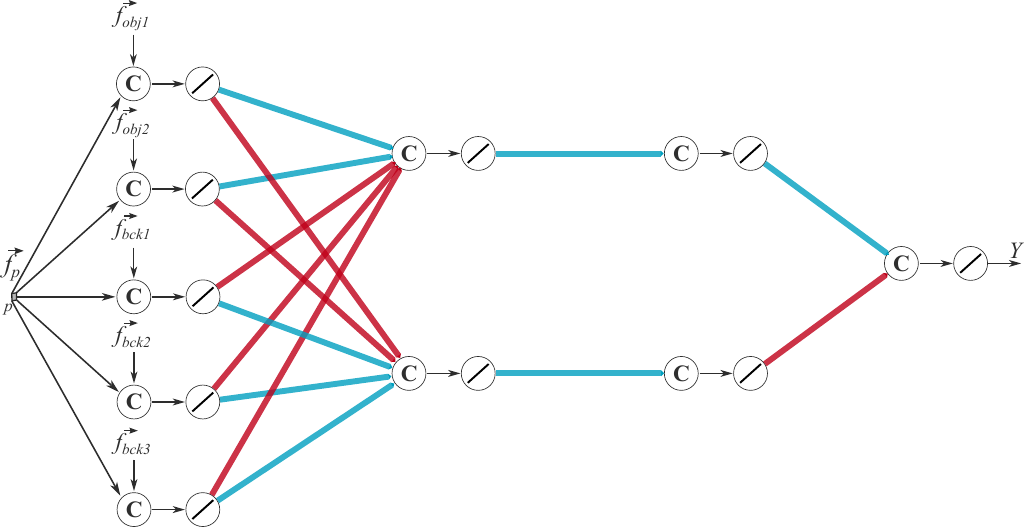}}
   \caption{(a): The 4-layer MMNN architecture is considered for the bean segmentation example. The first layer includes five multiset neurons with weights corresponding to the prototype and counter-prototype samples of the beans and other structures in the original image. Two neurons are used in the second and third layers, while the last layer incorporates a single neuron. The weight of the latter 5 neurons has been obtained by gradient ascent starting from the initial configuration shown in (b). The values of the thus optimized weights are indicated by the widths of the respective connections in the figure, with positive and negative values shown in blue and red, respectively. (b): The initial weight configuration adopted for the bean segmentation example. All weight elements are $1$, $-1$ or $0$. }\label{fig:Geral}
\end{figure}

\begin{figure}
  \centering
     \subfloat[]{\includegraphics[width=0.4\textwidth]{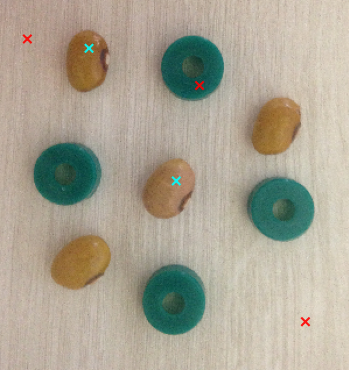}}
     \hspace{0.7cm}
     \subfloat[]{\fbox{\includegraphics[width=0.384\textwidth]{img_mask1.png}}}
     
     \vspace{0.9cm}
     
     \subfloat[]{\fbox{\includegraphics[width=0.384\textwidth]{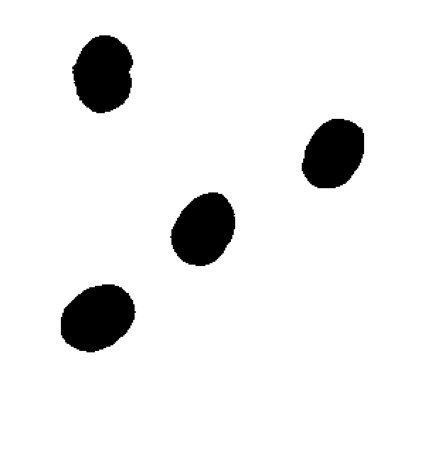}}}
     \hspace{0.7cm}
     \subfloat[]{\includegraphics[width=0.4\textwidth]{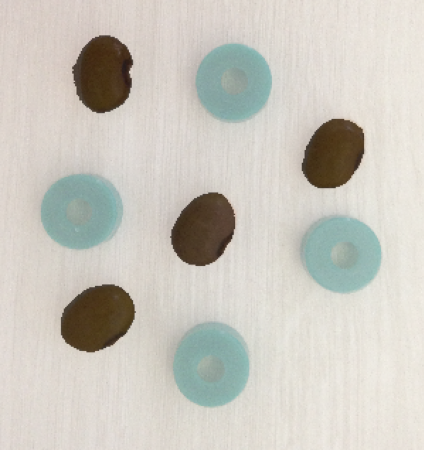}}
     
   \caption{(a): Indication of the prototype (cyan) and counter-prototype (red) samples respectively to the image from which the beans are to be segmented. (b): The respective gold standard. (c): The result of the beans segmentation using the four-layer MMNN architecture illustrated in Fig.~\ref{fig:Geral} with parameters $D=5$ and $r=3$ for the neurons in the first layer, $D=1$ for the neurons in the other layers, and optimized weights, leading to an overall accuracy of 98.67 \%. This result was obtained by thresholding the output of the MMNN with the value $T=0.46$. The numbers above the activation functions identify the respective outputs illustrated in Fig.~\ref{fig:Neurons}. (d): The superimposition of the segmented beans onto the original image. An improved segmentation of the beans has been obtained comparatively to the results shown in Figs.~\ref{fig:seg_beans} and~\ref{fig:img_3layars}, obtained by previously described architectures.}
  \label{fig:img_geral}
\end{figure}

In order to illustrate the potential of a more generic MMNN architecture, we revisit the task of segmenting the beans from the image in Figure~\ref{fig:img_geral}(a) by using the four-layer architecture illustrated in Figure~\ref{fig:Geral}. Linear activation functions are adopted for all multiset neurons in this architecture.

For simplicity's sake, the neurons in the first layer have been trained from respective prototype (cyan) and counter-prototypes (red) points taken from the several regions of the image, as illustrated in Figure~\ref{fig:img_geral}(a). Therefore, the neurons in the remainder layers 2 to 4 will then implement a non-linear combination of the indications about the several types of objects in the image in order to emphasize the beans. The respectively adopted gold standard, obtained by human inspection, is shown in Figure~\ref{fig:img_geral}(b)

The weights of the neurons from the second to the fourth layers are to be determined by using the gradient ascent methodology. More specifically, instead of considering a number of possible initial points, we start with the initial weight configuration shown in Figure~\ref{fig:Geral}(b).

This initial configuration assigns the two upper neurons in layers 2 and 3 to recognize the beans, while the lower neurons in those same layers are assigned to the other structures (i.e.~washers and background). No crossed links between the neurons in layers 2 and 3 are incorporated in the above initial configuration. The neuron in the fourth layer then compares the two respective inputs so as to emphasize the regions corresponding to the beans.

\begin{figure}
  \centering
     \includegraphics[width=1\textwidth]{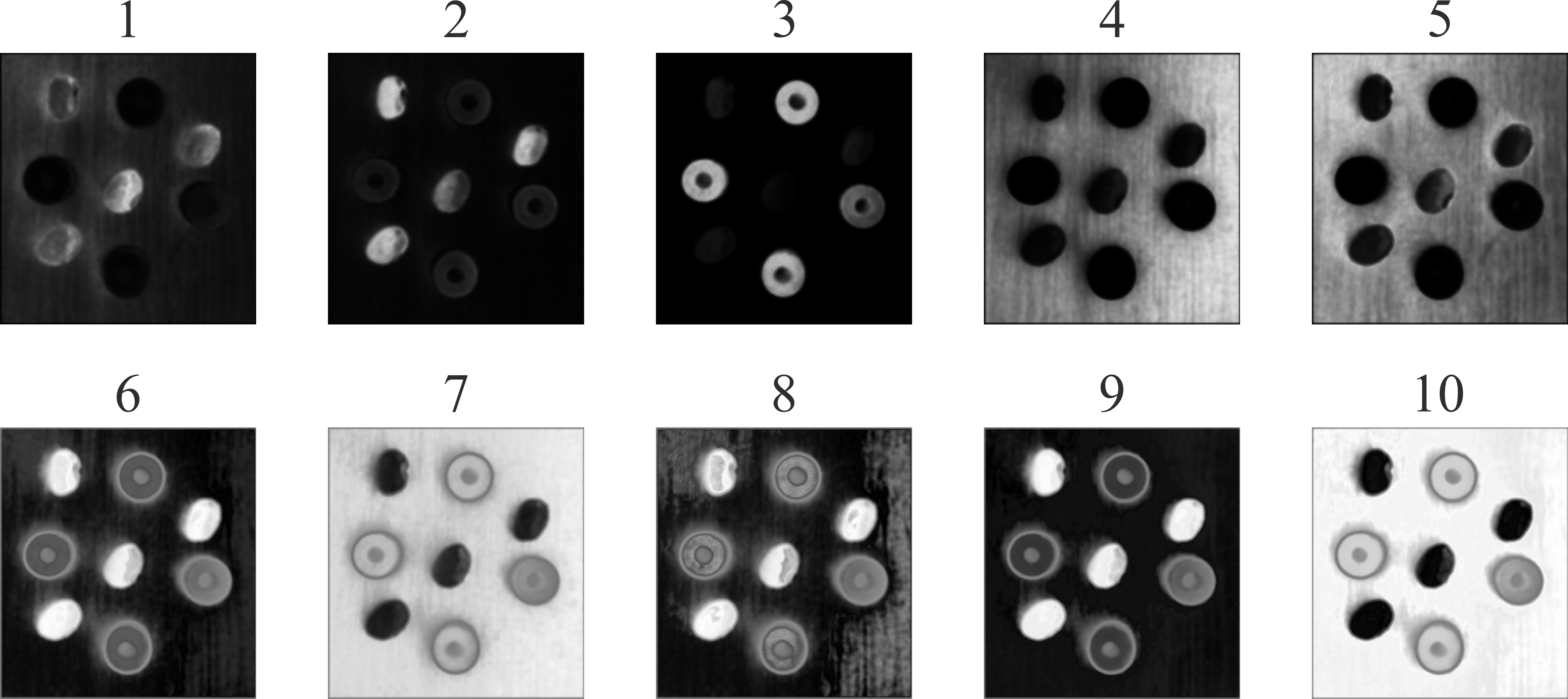}
   \caption{The intermediate segmentation results respective to each of the neurons in the adopted architecture, obtained for the optimized weights. Observe that the last image $10$, corresponding to the output $Y$ is complemented for compatibility with the adopted convention of representing the segmented objects in black.}\label{fig:Neurons}
\end{figure}

Figure~\ref{fig:Neurons} depicts the output of each of the multiset neurons resulting from the optimal weight configuration, identified by respective labels ranging from $1$ to $10$, with gray-level scale processing from dark to light gray levels. These images indicate the respective action of each of the multiset neurons, with the last image $10$ (the output $Y$) resulting in the beans being highlighted. 

Interestingly, the optimized weight configuration obtained by the gradient ascent starting at the above indicated initial weights resulted in substantially distinct, incorporating non-zero weights for the crossed links in layers 2 and 3. As can be inferred from Figure~\ref{fig:Neurons}, more specifically by comparing the outputs of neurons 7 and 10, the neurons in layers 2 and 3 implement enhancement of the contrast between the beans and other objects by making specific corrections, including subtraction of the upper left and lower right portions of the background.

When compared to the bean segmented images obtained by using the architectures described in Sections~\ref{sec:Two-Layer} and ~\ref{sec:Three-Layer}, the results in Figure~\ref{fig:img_geral} can be found to be more accurate, therefore indicating that the simultaneous training of the weights int the generic architecture in Figure~\ref{fig:Geral}(a) allowed respectively enhanced results.

\begin{figure}
  \centering
     \subfloat[]{\includegraphics[width=0.4\textwidth]{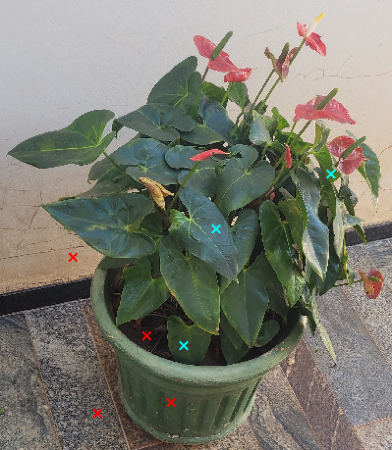}}
     \hspace{0.7cm}
     \subfloat[]{\fbox{\includegraphics[width=0.384\textwidth]{vaso_mask.png}}}
     
     \vspace{0.9cm}
     
     \subfloat[]{\fbox{\includegraphics[width=0.384\textwidth]{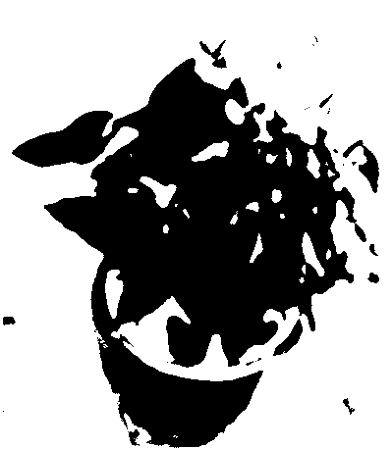}}}
     \hspace{0.7cm}
     \subfloat[]{\includegraphics[width=0.4\textwidth]{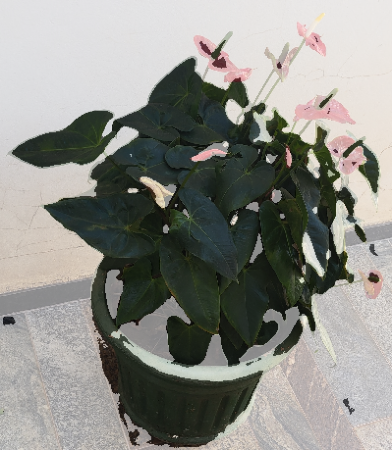}}
     
   \caption{(a): The three prototypes (cyan) and four counter-prototypes (red) were chosen as samples of the leaves and other structures. (b): The respective gold standard obtained by human inspection. (c): The segmentation results were achieved by using a generic four-layer MNN with 7, 3, 2, and 1 multiset neurons in the first, second, third, and fourth layers, leading to a balanced accuracy of $BA = 89.51 \%$. (d): The superimposition of the segmentation results and the original image.}
  \label{fig:vaseGeral}
\end{figure}

In order to further illustrate the potential of generic MMNN architectures, we revisit the segmentation of the leaves from the anthurium vase image addressed in terms of three neurons in Section~\ref{sec:Counter-Prototype}. A four-layer MNN architecture has been employed, containing 7, 3, 2, and 1 multiset neurons in the first, second, third, and fourth layers. All neurons have linear activation functions. The optimization was performed by gradient ascent while starting from an initial configuration similar to that used in the beans examples, using $D=3$, $r=4$, and $T=0.49$. A total of 3 prototype (cyan) and 4 counter-prototype (red) points have been chosen as shown in Figure~\ref{fig:vaseGeral}(a). The result is presented in Figure~\ref{fig:vaseGeral}(c), as well as its superimposition onto the original image shown in (d).

The respectively obtained balanced accuracy is higher than that previously obtained in Section~\ref{sec:Counter-Prototype} (86.47\% compared to 89.51 \%).  It can also be visually verified, by comparing with the respective gold standard in Figure~\ref{fig:vaseGeral}(b), that an enhanced segmentation of the anthurium leaves has been in fact achieved.

\section{Concluding Remarks}

Though effortless for humans and many other living beings, the ability to segment a given image into its constituent regions of potential interest constitutes a particularly challenging task to be computationally implemented, which has motivated continuing multidisciplinary research developments. In particular, image-related tasks, including segmentation and object recognition, have motivated and underlain several critically important developments in ANNs and deep learning.

While more traditional artificial neuronal models adopt the inner product as the initial stage, which constitutes a bi-linear operation, the possibility to replace this stage by a respective coincidence similarity operation, which is intrinsically non-linear, has been pursued more recently (e.g.~\cite{costa2023multiset, benatti2023two}) as a means of achieving potentially more strict and stable comparisons. Given that the coincidence similarity index can be understood as a multiset extension of the Jaccard index, the respectively obtained neurons have been called \emph{multiset neurons}.

The present work developed further the possibility of combining multiple multiset neurons into multi-layered multiset neuronal networks -- MNNs. While our focus was on utilizing MNNs for image segmentation and analysis (e.g.~\cite{costa2000shape,gonzales1987digital}), extensions to other types of data and applications should be readily obtained. More specifically, the initial layer has been associated with specific patterns to be recognized, being trained in terms of respective prototype and counter-prototype points. Additional neuronal layers are incorporated to enable pattern recognition of those identified in the first layer in a more generic and flexible manner.

While previous approaches to image segmentation by adopting multiset neurons have relied mostly on the coincidence index for comparing two vectors with positive entries, the possibility to use counter-prototype points adopted in the present work implied the use of multiset neurons capable of comparing vectors with real entries. As illustrated in the present work, the use of combined prototype and counter-prototype allowed the segmentation of regions in intricate images (e.g.~anthurium vase) to be performed with impressive accuracy and effectiveness while using only one prototype and one counter-prototype points, therefore involving only two multiset neurons.

The main challenge while combining multiset neurons concerns the methodology to be employed for the determination of the respective weights, which constituted the main subject of the present work. The task of determining the weights of the multiset neurons in subsequent layers has been approached here in terms of a gradient ascent optimization methodology taking place on a respective balanced accuracy landscape (a scalar field). Several interesting results have been obtained that reinforce the potential of MNNs for effective generic applications in pattern recognition problems.

The first interesting finding reported here shows that achieving accurate segmentation of particular regions can often be accomplished by utilizing one prototype and corresponding counter-prototype points. This makes it possible to visualize not only the balanced accuracy landscape but also the trajectories formed by the gradient ascent optimizations carried out on these landscapes. Additional examples included two approaches to joint segmentation of two types of objects by using a 3-layer MMN. These methods involve separated and simultaneous training of the neurons at the second and third layers.

It has been shown that relatively simple and smooth landscapes are obtained, which contributed to the effective performance of gradient ascent optimization, which led to adequate solutions in every considered situation. However, as these landscapes have been found to typically incorporate more than a single attraction basin, it becomes important to perform the gradient optimization while considering several initial points, so that the trajectory leading to the highest balanced accuracy can be chosen as possible (but not certain) global maximum.

In order to complement our presentation and study, a completely generic MMNN architecture has been described which involves any number of layers and neurons per layer. The potential of this approach has been successfully illustrated respectively to achieving improved segmentation results when compared to the other approaches previously described.

The reported results open up numerous avenues for further research in pattern recognition, ANNs, deep learning, and Artificial Intelligence. Among these possibilities, it could be particularly promising to explore substantially larger networks having more layers, thereby extending the current study to deep learning architectures. Designing and implementing deep learning architectures using multiset neurons is particularly interesting, allowing for the combination of multiset neurons and traditional neuronal models within the same network, using the inner product as the initial stage. Other possibilities include considering optimization approaches such as simulated annealing as a means of identifying the weights of the multiset neurons involved. Adapting the backpropagation method (e.g.~\cite{haykin1998neural,wythoff1993backpropagation,goh1995back}) for training MNNs is of special interest. Implementing MNNs in customized hardware (both discrete and integrated) is yet another fascinating possibility, motivated by the simple implementation of multiset operations in analog hardware~\cite{da2021signal}.

\section*{Acknowledgments}
Luciano da F. Costa thanks CNPq (grant no. 307085/2018-0) and FAPESP (grant 15/22308-2).

\bibliography{ref}

\begin{thebibliography}{10}

\bibitem{goldstein2021sensation}
E.~B. Goldstein and L.~Cacciamani.
\newblock {\em Sensation and perception}.
\newblock Cengage Learning, 2021.

\bibitem{haykin1998neural}
S.~Haykin.
\newblock {\em Neural networks: a comprehensive foundation}.
\newblock Prentice Hall PTR, 1998.

\bibitem{kohonen1990self}
T.~Kohonen.
\newblock The self-organizing map.
\newblock {\em Proceedings of the IEEE}, 78(9):1464--1480, 1990.

\bibitem{lecun2015deep}
Y.~LeCun, Y.~Bengio, and G.~Hinton.
\newblock Deep learning.
\newblock {\em Nature}, 521(7553):436--444, 2015.

\bibitem{krizhevsky2017imagenet}
A.~Krizhevsky, I.~Sutskever, and G.~E. Hinton.
\newblock Imagenet classification with deep convolutional neural networks.
\newblock {\em Communications of the ACM}, 60(6):84--90, 2017.

\bibitem{pouyanfar2018survey}
S.~Pouyanfar, S.~Sadiq, Y.~Yan, H.~Tian, Y.~Tao, M.~P. Reyes, M.-L. Shyu, S.-C.
  Chen, and S.~S. Iyengar.
\newblock A survey on deep learning: Algorithms, techniques, and applications.
\newblock {\em ACM Computing Surveys (CSUR)}, 51(5):1--36, 2018.

\bibitem{costa2023multiset}
L.~da~F. Costa.
\newblock Multiset neurons.
\newblock {\em Physica A: Statistical Mechanics and its Applications},
  609:128318, 2023.

\bibitem{costa2021multiset}
L.~da~F.~Costa.
\newblock {Multiset Self-Organizing Map--MSSOM}.
\newblock
  \url{https://www.researchgate.net/publication/356170912_Multiset_Self-Organizing_Map_-_MSSOM},
  2021.

\bibitem{benatti2023two}
A.~Benatti and L.~da~F. Costa.
\newblock Two approaches to supervised image segmentation.
\newblock {\em arXiv preprint arXiv:2307.10123}, 2023.

\bibitem{costa2022on}
L.~da~F.~Costa.
\newblock On similarity.
\newblock {\em Physica A: Statistical Mechanics and its Applications},
  599:127456, 2022.

\bibitem{da2022abrief}
L.~da~F.~Costa.
\newblock A brief guide to the coincidence similarity and its applications.
\newblock {\em ResearchGate}, 2022.

\bibitem{da2022supervised}
L.~da~F. Costa.
\newblock Supervised image segmentation by using multiset similarity neurons.
\newblock
  \url{https://www.researchgate.net/publication/360842447_Supervised_Image_Segmentation_by_Using_Multiset_Similarity_Neurons},
  2022.

\bibitem{Jaccard1901distribution}
P.~Jaccard.
\newblock Distribution de la flore alpine dans le bassin des dranses et dans
  quelques r\'egions voisines.
\newblock {\em Bulletin de la Soci\'et\'e vaudoise des sciences naturelles},
  37:241--272, 1901.

\bibitem{Jac:wiki}
Wikipedia.
\newblock Jaccard index, 2021.
\newblock \url{https://en.wikipedia.org/wiki/Jaccard_index}. [Online; accessed
  10-Oct-2021].

\bibitem{leydesdorff2008on}
L.~Leydesdorff.
\newblock On the normalization and visualization of author co-citation data:
  Salton's cosine versus the jaccard index.
\newblock {\em Journal of the American Society for Information Science and
  Technology}, 59(1):77--85, 2008.

\bibitem{da2021further}
L.~da~F.~Costa.
\newblock Further generalizations of the {J}accard index.
\newblock
  \url{https://www.researchgate.net/publication/355381945_Further_Generalizations_of_the_Jaccard_Index},
  2021.

\bibitem{vijaymeena2016a}
M.~K. Vijaymeena and K.~Kavitha.
\newblock A survey on similarity measures in text mining.
\newblock {\em Machine Learning and Applications: An International Journal},
  3(2):19--28, 2016.

\bibitem{CostaCCompl}
L.~da~F.~Costa.
\newblock Coincidence complex networks.
\newblock {\em J. Phys. Complex.}, 3:015012, 2022.

\bibitem{egmont2002image}
M.~Egmont-Petersen, D.~de~Ridder, and H.~Handels.
\newblock Image processing with neural networks—a review.
\newblock {\em Pattern Recognition}, 35(10):2279--2301, 2002.

\bibitem{alom2018nuclei}
Md~Z. Alom, M.~Hasan, C.~Yakopcic, T.~M. Taha, and V.~K. Asari.
\newblock Nuclei segmentation with recurrent residual convolutional neural
  networks based u-net (r2u-net).
\newblock In {\em NAECON 2018-IEEE National Aerospace and Electronics
  Conference}, pages 228--233. IEEE, 2018.

\bibitem{minaee2021image}
S.~Minaee, Y.~Boykov, F.~Porikli, A.~Plaza, N.~Kehtarnavaz, and D.~Terzopoulos.
\newblock Image segmentation using deep learning: A survey.
\newblock {\em IEEE Transactions on Pattern Analysis and Machine Intelligence},
  44(7):3523--3542, 2021.

\bibitem{cheng2001color}
H.-D. Cheng, X.~H. Jiang, Y.~Sun, and J.~Wang.
\newblock Color image segmentation: advances and prospects.
\newblock {\em Pattern Recognition}, 34(12):2259--2281, 2001.

\bibitem{pham2000current}
D.~L. Pham, C.~Xu, and J.~L. Prince.
\newblock Current methods in medical image segmentation.
\newblock {\em Annual Review of Biomedical Engineering}, 2(1):315--337, 2000.

\bibitem{du2019gradient}
S.~Du, J.~Lee, H.~Li, L.~Wang, and X.~Zhai.
\newblock Gradient descent finds global minima of deep neural networks.
\newblock In {\em International Conference on Machine Learning}, pages
  1675--1685. PMLR, 2019.

\bibitem{ruder2016overview}
S.~Ruder.
\newblock An overview of gradient descent optimization algorithms.
\newblock {\em arXiv preprint arXiv:1609.04747}, 2016.

\bibitem{da2021multiset}
L.~da~F. Costa.
\newblock Multiset-based image segmentation.
\newblock
  \url{https://www.researchgate.net/publication/356563786_Multiset-Based_Image_Segmentation},
  2021.

\bibitem{da2022multisets}
L.~da~F. Costa.
\newblock Multisets.
\newblock \url{https://www.researchgate.net/publication/355437006_Multisets},
  2022.

\bibitem{blizard1989multiset}
W.~D. Blizard et~al.
\newblock Multiset theory.
\newblock {\em Notre Dame Journal of Formal Logic}, 30(1):36--66, 1989.

\bibitem{blizard1991development}
W.~D. Blizard.
\newblock The development of multiset theory.
\newblock {\em Modern Logic}, 1:319--352, 1991.

\bibitem{lettvin1959frog}
J.~Y. Lettvin, H.~R. Maturana, W.~S. McCulloch, and W.~H. Pitts.
\newblock What the frog's eye tells the frog's brain.
\newblock {\em Proceedings of the IRE}, 47(11):1940--1951, 1959.

\bibitem{gonzales1987digital}
R.~C. Gonzales and P.~Wintz.
\newblock {\em Digital image processing}.
\newblock Addison-Wesley Longman Publishing Co., Inc., 1987.

\bibitem{costa2000shape}
L.~da~F. Costa and R.~M. Cesar~Jr.
\newblock {\em Shape analysis and classification: theory and practice}.
\newblock CRC Press, Inc., 2000.

\bibitem{garcia2009index}
V.~Garc{\'\i}a, R.~A. Mollineda, and J.~S. S{\'a}nchez.
\newblock Index of balanced accuracy: A performance measure for skewed class
  distributions.
\newblock In {\em Iberian conference on pattern recognition and image
  analysis}, pages 441--448. Springer, 2009.

\bibitem{press2007numerical}
W.~H. Press.
\newblock {\em Numerical recipes 3rd edition: The art of scientific computing}.
\newblock Cambridge University Press, 2007.

\bibitem{wythoff1993backpropagation}
B.~J. Wythoff.
\newblock Backpropagation neural networks: a tutorial.
\newblock {\em Chemometrics and Intelligent Laboratory Systems},
  18(2):115--155, 1993.

\bibitem{goh1995back}
A.~T.~C. Goh.
\newblock Back-propagation neural networks for modeling complex systems.
\newblock {\em Artificial Intelligence in Engineering}, 9(3):143--151, 1995.

\bibitem{da2021signal}
L.~da~F. Costa.
\newblock Multiset signal processing and electronics.
\newblock
  \url{https://www.researchgate.net/profile/Luciano-Da-F-Costa/research}, 2022.

\end{thebibliography}
\bibliographystyle{unsrt}

\end{document}